\pdfoutput=1

\documentclass[11pt]{article}

\usepackage[final]{acl}

\usepackage{times}
\usepackage{latexsym}

\usepackage[T1]{fontenc}

\usepackage[utf8]{inputenc}

\usepackage{microtype}

\usepackage{inconsolata}

\usepackage{graphicx}

\usepackage{multirow}

\title{Retrieval-Augmented Simulacra:\\ Generative Agents for Up-to-date and Knowledge-Adaptive Simulations}

\author{Hikaru Shimadzu$^1$ \ \  Takehito Utsuro$^2$ \ \ Daisuke Kitayama$^1$  \\
  $^1${School of Informatics, Kogakuin University} \\
  $^2${Deg. Prog. Sys.\&Inf. Eng., Grad. Sch. Sci.\&Tech., University of Tsukuba} \\
  \texttt{\{j323402\_@\_ns, kitayama\_@\_cc\}.kogakuin.ac.jp}, \  \texttt{utsuro\_@\_iit.tsukuba.ac.jp}
}

\begin{document}
\maketitle
\begin{abstract}
In the 2023 edition of the White Paper on Information and Communications, it is estimated that the population of social networking services in Japan will exceed 100 million by 2022, and the influence of social networking services in Japan is growing significantly. In addition, marketing using SNS and research on the propagation of emotions and information on SNS are being actively conducted, creating the need for a system for predicting trends in SNS interactions. We have already created a system that simulates the behavior of various communities on SNS by building a virtual SNS environment in which agents post and reply to each other in a chat community created by agents using a LLMs. In this paper, we evaluate the impact of the search extension generation mechanism used to create posts and replies in a virtual SNS environment using a simulation system on the ability to generate posts and replies. As a result of the evaluation, we confirmed that the proposed search extension generation mechanism, which mimics human search behavior, generates the most natural exchange.
\end{abstract}

\section{Introduction}

Social networking services (SNS) have been integrated into people's lives as a means of casual communication. In Japan, the trend is particularly pronounced and, According to the 2023 Information and Communications White Paper \footnote{https://www.soumu.go.jp/johotsusintokei\\/whitepaper/en/r05/html/nd247100.html} , the number of social media users in Japan in 2022 is estimated to be over 100 million. In addition, marketing and demand surveys using SNS are also conducted by a great number of companies, indicating that society as a whole uses SNS frequently. 

One of the characteristics of SNSs is the difficulty of predicting trends, since SNSs allow a large number of unspecified users to interact in a variety of ways.This difficulty also makes it difficult for companies to predict the effects of marketing and other activities and to analyze them as research subjects.

\begin{figure}[t]
    \centering
    \includegraphics[width=0.99\linewidth]{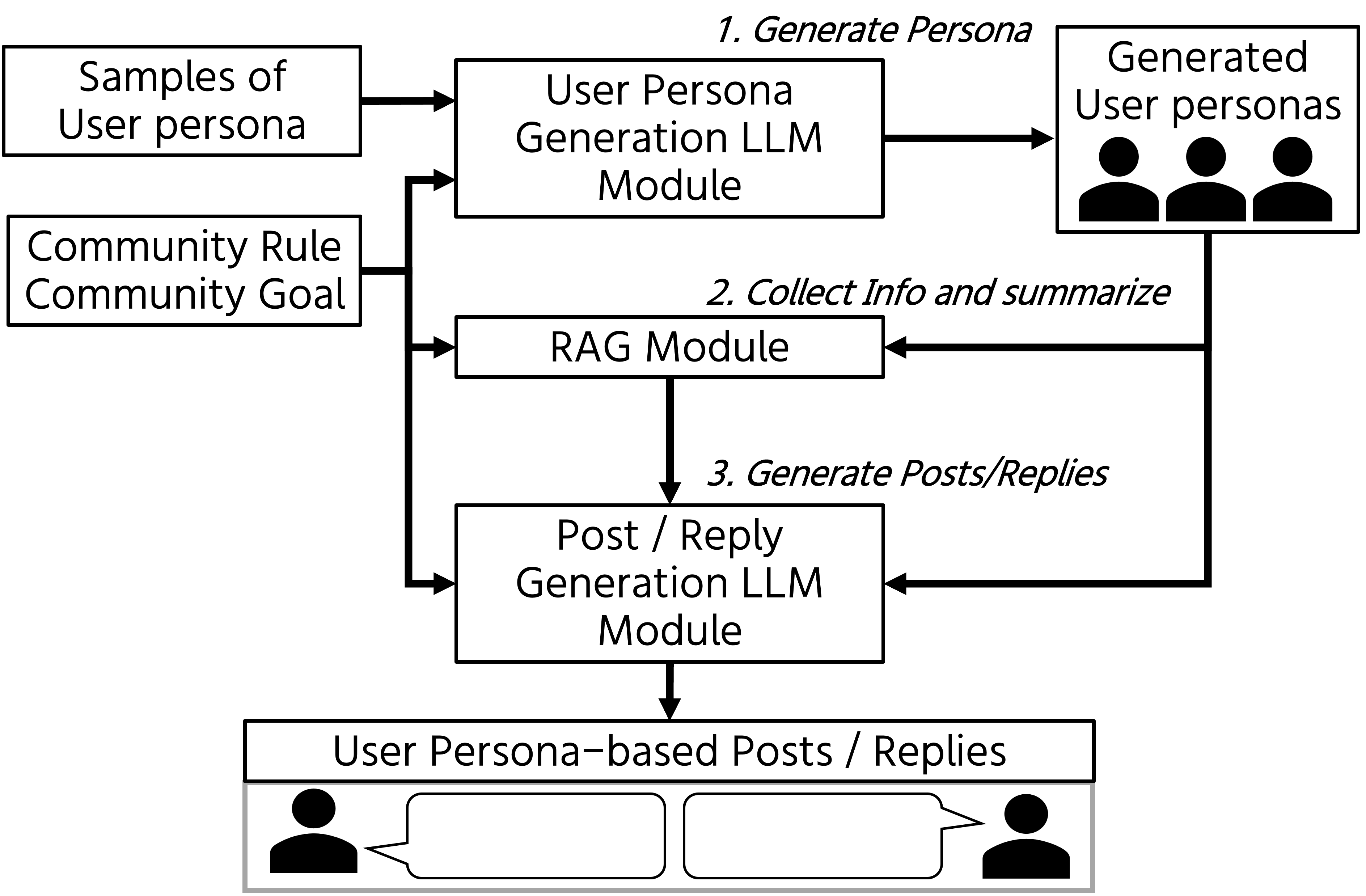}
    \caption{Ovewview of system}
    \label{fig:システム概要}
\end{figure}

So we proposed a system that simulates user interactions and mimics and predicts interactions using multiple chat agents that operate using LLMs. In this system, multiple chat agents that create posts and replies using LLMs are created, and by having them exchange posts and replies with each other, we simulate text-based communication on social networking sites as if it were being carried out by humans. Figure \ref{fig:システム概要} shows an overview of the system proposed in this study.

As shown in Figure \ref{fig:システム概要}, we have already created a system that simulates a SNS by having a chat agent that generates responses according to a specific persona behave as a SNS user belonging to a community with a certain purpose and rules. The system enables agents to generate realistic and accurate posts and replies by using information collected from the Web, and also enables simulations of the latest topics.


By using this system to simulate a particular topic or post, it is possible to predict how that topic or post will be interpreted by a given community. This prediction can be used to predict responses to marketing and advertising, and can also be used as a target for analysis of emotion propagation and information transfer by analyzing agent interactions over time.

In this paper, we investigate how much the RAG(Retrieval Augmented Generation) mechanism in the created SNS simulation system affects the simulation capability, and also investigate how the proposed method, RAG mechanism considering human search behavior, affects the simulation.

\section{Related Work}
\subsection{Reproduction of Communication and Related Phenomena Among Humans Using LLM}

\citet{hua2024warpeacewaragentlarge} constructed a diplomatic simulation system called ``WarAgent'' that uses conversations between the agents with LLM that have been given background information on political systems, military strength, resources, etc., and showed that LLMs that have been given appropriate background information can construct diplomatic relations that are close to historical facts.

\citet{ohagi2024polarizationautonomousgenerativeai} have also created a discussion simulation system using multiple agents with LLMs, and is investigating how the echo chamber phenomenon changes the opinions of each agent. Ohagi has shown that the echo chamber phenomenon also occurs in LLMs when the system is adjusted to have more discussions with agents of the same opinion.

Elsewhere, \citet{pmlr-v202-aher23a} have conducted psychological experiments with human subjects on LLMs to investigate whether they can replicate human results. A study by Aher et al. reported that LLMs gave comparable responses and experimental results to several psychological and psycholinguistic experiments, such as an ultimatum game, reading cul-de-sac sentences, and a Milgram experiment, to humans.

These studies by Hua et al., Ohagi, and Aher et al. suggest that LLMs have the ability to reproduce and mimic human communication and its associated phenomena.

\subsection{Simulation of Behavior Using LLM}

\citet{10.1145/3526113.3545616} has also conducted a simulation of interactions on social networking sites using chat agents with LLMs, Our research is based on the work of Park et al. In addition to the method proposed by Park et al., our research uses RAG to enable simulations on the latest topics. Park et al. only evaluate the posts and replies generated by the agents for each individual post or reply. In addition to these, our research evaluates the series of interactions between posts and replies. This is to evaluate the quality of the simulation itself.

\citet{park2023generativeagentsinteractivesimulacra} have also simulated real-world human behavior with a person agent controlled by a LLMs. Park et al. extended the SNS interaction simulation described above to reality, mimicking and simulating human behavior through interrelationships between human agents who perform behaviors such as eating and sleeping as humans do. The simulation differs from the present study in that the subject of the simulation is physical behavior in the real world.

\citet{gao2023s3socialnetworksimulationlarge} has also implemented a social network simulation system called ``$S^3$'' , similar to our research. They are similar in that they simulate user interactions on social networks, but Chen et al. differ in that they simulate the propagation of information and emotions within interactions, whereas our research simulates the interactions themselves.

\section{Proposed Method}

\subsection{Assumptions of The Proposed Method}
In the SNS exchange simulation system handled in this research, SNS is abstracted and considered as a service with the following functions.
\begin{itemize}
    \item Users can post or reply any text.
    \item The user can add a reply to the post.
    \item All users can view any postings and replies.
\end{itemize}

In many cases, real-life SNSs are designed to allow users to communicate in multiple ways, including ``reposting'' and reaction functions such as ``like'' or ``favorite'' in addition to these. In this study, we focused on communication centered on the exchange of opinions through text.

\subsection{Overview of The Proposed Method}

In our research, we propose a system that reproduces interactions between users on SNS by means of multi-agent simulation using multiple agents that generate posts using LLMs. In this system, LLMs play the roles of multiple users on SNS, and they imitate interactions between users by exchanging text-based posts and replies.

As shown in Figure \ref{fig:システム概要}, this SNS interaction simulation system is composed of two modules that use LLM: ``User persona generation LLM'' and ``Post/reply generation LLM''. This module structure is inherited from ``Social Simulacra'' by Park et al. However, in our research, we have added a summary mechanism for collecting information for Retrieval Augmented Generation \cite{10.5555/3495724.3496517} to the posting and reply module.

The Table \ref{tab:シミュレーション段階} shows the stages of the simulation and an overview of each stage. When the simulation starts, the first step is to generate users. By providing a sample set of community rules, topics, and expected users, other users that could exist in the community are generated. Next, a LLMs role-plays the users and creates posts to simulate user interaction.The generated posts are generated by RAG using the latest information available on the Web in order to make the simulation more realistic. The details of each stage are explained in the following sections in the order of the stages. All prompts and other information used at each stage are listed in the Figure \ref{tab:ペルソナ生成プロンプト} through \ref{tab:返信生成プロンプト} in Appendix \ref{sec:提案手法付録}.

\begin{table*}[t]
    \centering
    \small
    \begin{tabular}{|c|c|}
    \hline
    Stage & Overviws\\
    \hline
    \hline
    Generate user personas & \begin{tabular}{c}Using the few-shot prompt with the given sample user persona, the agents \\generates a user persona with features associated with the community goal.\end{tabular}\\
    \hline
    Information collection \& summarize & \begin{tabular}{c}Based on the community settings, the agents searches the web for \\related informations and prepares to use informations to generate the post.\end{tabular}\\
    \hline
    Generate posts \& replies & \begin{tabular}{c}Based on the community settings, the agents generates posts and replies \\using the information retrieved in the previous step.\end{tabular}\\
    \hline
    \end{tabular}
    \caption{Simulation Stages and Overview}
    \label{tab:シミュレーション段階}
\end{table*}

\subsection{Stage 1: Generate Personas\label{sec:ペルソナ生成}}

In this study, we define a user persona as a short sentence that describes a characteristic of a user. LLMs are known to change the quality and characteristics of the responses they generate when given specific job titles and characteristics \cite{xu2023expertpromptinginstructinglargelanguage}. By using this, it is possible to reproduce a community environment on a social networking service where users with various characteristics are mixed.

\begin{figure}[t]
    \centering
    \includegraphics[width=0.9\linewidth]{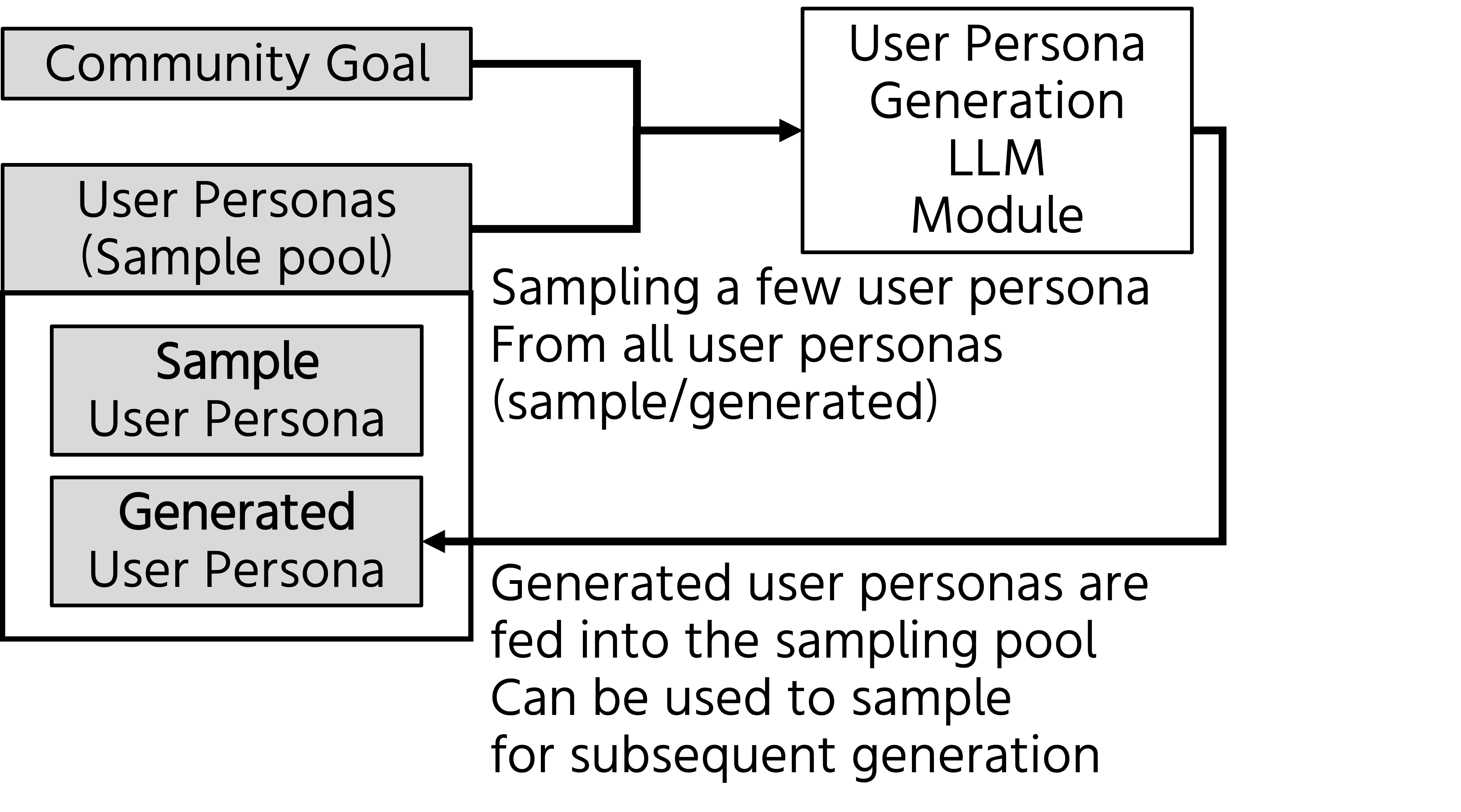}
    \caption{User persona generation module}
    \label{fig:ユーザペルソナ概要図}
\end{figure}

In generating user personas, we sample up to 10 given or previously generated user personas and create Few-shot prompts, as shown in Figure \ref{fig:ユーザペルソナ概要図}. The prompt also includes the community goal.

As an example of a user persona, a "disaster prevention engineering expert" or a "corporate disaster prevention manager" can be assumed when running a simulation that aims to "exchange opinions on disaster prevention.

Note that three parameters, ``attention'', ``range'' and ``depth'', are given at the same time when generating user personas. Of these, ``attention'' and ``range'' are random values rounded to integers according to a normal distribution: $\mathcal{N}(5,1)$. In this case, they are processed so that the values range from 1 to 10. The ``depth'' is given a random value according to the Erlang distribution:$Er(3,1)$ and rounded to an integer, taking the range from 0 to 6.The details of these parameters are explained in section \ref{sec:RAG}.

\subsection{Stage 2: Information Collection \& Summarize\label{sec:RAG}}

A common characteristic of LLMs is that they may return an incorrect response, ``Halsination'' on topics about which they have no knowledge. halsination occurs frequently on topics that have since been modeled, on topics related to specific places, people, and organizations, and on specialized topics that require advanced knowledge. Halcination can lead to the generation of posts or replies that deviate excessively from the facts, thus undermining the reliability of the simulation results. In order to enable the simulation of exchanges on topics that are up-to-date or require specific knowledge, this system will perform RAGs using information on the Web. This enables us to reproduce a real SNS environment where users can freely search for information, and to reproduce and simulate natural discussions and exchanges on any topic.

\begin{figure*}[t]
    \centering
    \includegraphics[width=0.8\linewidth]{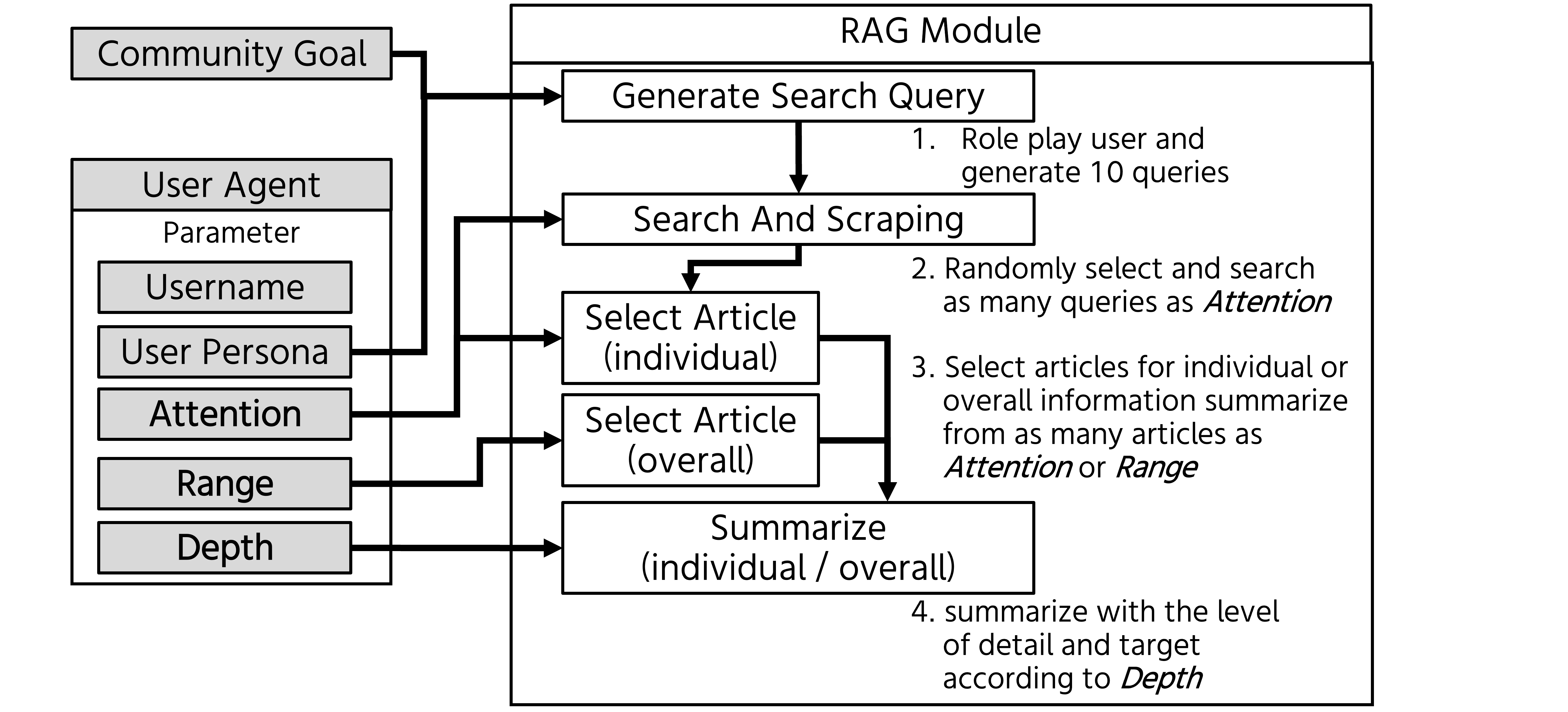}
    \caption{RAG module}
    \label{fig:RAG概要図}
\end{figure*}

When conducting research on a topic, we assumed the search behavior of collecting both general information about the topic and detailed information about matters of particular interest to the actor. To replicate this, when a user persona is generated, three parameters are given to control the treatment of information for RAGs: ``attention'', ``range'', and ``depth''. Figure \ref{fig:RAG概要図} shows an image of the RAG module in operation.

First, The module generates a 10 search query. Query generation is based on the community goals and the user personas, and in the case of replies, the thread content is also used.

Next, among the 10 search queries generated in the previous step, randomly select a number of queries equal to the ``attention'' value given to the user persona, and scrape top 10 articles in search results for each query. In this paper, NHK NEWS WEB\footnote{https://www3.nhk.or.jp/news/} was used as the source of scraping.

The scraped articles are summarized using LexRank\cite{10.5555/1622487.1622501} based on the ``depth'' value and length of article $l$. The amount of text after summarization and target of summarization is determined as shown in Table \ref{tab:要約文長さ} in Appendix \ref{sec:提案手法付録}. The same number of articles as the ``range'' is used to capture the overall information, and the same number of articles as the ``attention'' is used to capture the individual information. This mechanism is a mimicry of the human search behavior mentioned above.

\subsection{Stage 3: Generate Posts \& Replies\label{sec:投稿生成}}

Generate posts and replies with prompts containing user persona, community goals and rules, and collected and summarized information. 

First, a certain number of user personas are selected from all generated user personas to create a top-level post. A top-level post is an independent post that serves as a starting point for discussion, etc., and does not have a reply target. When a top-level post is created, a new thread with that post is created. By creating top-level posts using a certain number of randomly sampled personas, a thread is prepared that is selected as a reply target in the later stage of reply generation. 

Once the generation of top-level posts is complete, replies are generated. A reply is a post made in response to a top-level post or reply, and is posted as an addition to an existing thread. Replies are generated by all user personas to a randomly selected thread. If the number of posts/replies in a thread is greater than 10, another thread will be reselected.

Top-level posts and replies are managed in ``threads'', where the oldest post is placed at the top of the thread and the newest post is placed at the bottom of the thread. Every thread has one top-level post and zero or more replies. 

\section{Experiment}

In this study, we evaluate post/reply and threads to investigate how the RAG mechanism built into the simulation system affects the simulation system.For each generation, we used rinna-llama-3-youko-70b-instruct-Q4\_K\_M as the model, as well as a set of pre-created user personas.

The following sections provide details on the evaluation methods used for each evaluation target and the experiments conducted in this paper.

\subsection{Evaluation Methodology for Posts and Threads}
In the evaluation of posts and threads, in order to examine the impact of the RAG mechanism on simulations, simulations were run with the following three conditions, and the generated posts and replies and threads were evaluated.

\begin{itemize}
    \item Simulation without RAG mechanism
    \item Simulation with simple RAG mechanism
    \item Simulation with the advanced RAG mechanism (proposed method)
\end{itemize}

In the simulation without the RAG mechanism, the simulation is run without any information from the Web when generating posts and replies.

In the simulation with the simple RAG mechanism, the article search was performed with a search query created using LLM at the beginning of the simulation, and the top three search results were summarized and given.With this mechanism, the information used for post and reply generation is the same for all agents.

For simulations with the advanced RAG mechanism, the simulations are run using the proposed method described in section \ref{sec:RAG}.

For the post evaluation, the purpose of the community interaction(community goal), the rules of the community(community rule), the description of the person who created the post or reply(user persona), and the text of the generated posts or replies were presented and subjectively evaluated six times each by different subjects. When evaluating them, we instruct the authors to evaluate only their content, ignoring the context before and after it. The evaluation items are as follows. The evaluation included 28 top-level posts and 110 replies for each scenario and condition.

\begin{enumerate}
    \item Is there any grammatical errors in the content of the post?
    \item Is the content of the post appropriate for the community goals?
    \item Is the content of the post adhere to the community rules?
    \item Is the content of the post as imagined by the persona used to generate the post?
    \item Is the content of the post natural as a post on an internet forum with that community goal?
\end{enumerate}

In the thread evaluation, the community goal, community rules, and all posts in the thread (top-level posts and replies) are presented, and three different subjects are asked to subjectively evaluate the following items for the interactions taking place in the thread. Threads for evaluation were those in which one or more replies existed, and threads with only top-level posts were excluded.

\begin{enumerate}
    \item Is there any grammatical errors in the content of the exchange?
    \item Is the content of the exchange is appropriate for the community goal?
    \item Is the content of the exchange in compliance with the community rules?
    \item Is the exchange natural as an exchange on an internet forum with that community goal?
\end{enumerate}

\subsection{Conditions for Running The Simulation}

Two simulation scenarios, ``Bank of Japan Discussion'' and ``Otani Chat'', were assumed in the simulation runs. The community goals and rules set for each scenario are shown in below. These scenarios were set up for topics that require up-to-date or accurate information, for which conventional methods without RAGs would not be able to run valid simulations.

\begin{itemize}
    \item To discuss the Bank of Japan's monetary policy
    \begin{description}
        \item[Goal] To discuss the Bank of Japan's monetary policy.
        \item[Rule] Use courteous language in discussions.
    \end{description}
    \item Chat about Shohei Ohtani's success
    \begin{description}
        \item[Goal] To chat about Shohei Ohtani's activities.
        \item[Rule] Avoid topics not related to Shohei Ohtani.
    \end{description}
\end{itemize}

\begin{table*}[t]
    \centering
    \small
    \begin{tabular}{|c|c||c|c|c|c|}
        \hline
        Scenario                  & Conditions    & Compatible & Somewhat compatible & Somewhat incompatible & Incompatible \\
        \hline
        \hline
        \multirow{3}{*}{BOJ Discussion} & Without RAG & 288 & 103 & 21 & 2\\
                              & Simple RAG & 287 & 90 & 25 & 12\\
                              & Adv. RAG & 295 & 94 & 21 & 4\\
        \hline
        \multirow{3}{*}{Otani Chat} & Without RAG & 311 & 82 & 20 & 1\\
                              & Simple RAG & 324 & 73 & 16 & 1\\
                              & Adv. RAG & 283 & 96 & 31 & 4\\
        \hline
    \end{tabular}
     \caption{Results of the evaluation of the conformity of posts and replies to the community goals}
    \label{tab:投稿目標適合性評価}
\end{table*}

\begin{table*}[t]
    \centering
    \small
    \begin{tabular}{|c|c||c|c|c|c|}
        \hline
        Scenario                  & Conditions    & Compatible & Somewhat compatible & Somewhat incompatible & Incompatible \\
        \hline
        \hline
        \multirow{3}{*}{BOJ Discussion} & Without RAG & 332 & 71 & 10 & 1\\
                              & Simple RAG & 315 & 79 & 16 & 4\\
                              & Adv. RAG & 330 & 70 & 9 & 5\\
        \hline
        \multirow{3}{*}{Otani Chat} & Without RAG & 322 & 77 & 15 & 0\\
                              & Simple RAG & 349 & 52 & 10 & 3\\
                              & Adv. RAG & 270 & 84 & 40 & 20\\
        \hline
    \end{tabular}
    \caption{Results of the evaluation of the conformity of posts and replies to the community rules}
    \label{tab:投稿ルール適合性評価}
\end{table*}

\begin{table*}[t]
    \centering
    \small
    \begin{tabular}{|c|c||c|c|c|c|}
        \hline
        Scenario                 & Conditions    & natural & somewhat natural & somewhat unnatural & unnatural\\
        \hline
        \hline
        \multirow{3}{*}{BOJ Discussion} & without RAG  & 219 & 168 & 25 & 2\\
                              & Simple RAG & 218 & 137 & 45 & 14\\
                              & Adv. RAG & 228 & 139 & 41 & 6\\
        \hline
        \multirow{3}{*}{Otani Chat} & without RAG  & 245 & 140 & 22 & 7\\
                              & Simple RAG & 256 & 118 & 31 & 9\\
                              & Adv. RAG & 219 & 139 & 41 & 15\\
        \hline
    \end{tabular}
    \caption{Results of evaluation of the naturalness of the posts and replies}
    \label{tab:投稿自然さ評価}
\end{table*}

\begin{table*}[t]
    \centering
    \small
    \begin{tabular}{|c|c||c|c|c|c|c|}
        \hline
        Scenario                  & Conditions    &\begin{tabular}{c}Compatible\end{tabular} & \begin{tabular}{c}Somewhat\\compatible\end{tabular} & \begin{tabular}{c}Somewhat\\incompatible\end{tabular} & Incompatible & Eval. total\\
        \hline
        \hline
        \multirow{3}{*}{\begin{tabular}{c}BOJ\\Discussion\end{tabular}} & Without RAG & 66.67\% & 25.33\% & 6.67\% & 1.33\% & 75\\
                              & Simple RAG & 50.00\% & 34.62\% & 10.26\% & 5.13\% & 78\\
                              & Adv. RAG & 69.14\% & 22.22\% & 8.64\% & 0\% & 81\\
        \hline
        \multirow{3}{*}{Otani Chat} & Without RAG & 72.84\% & 23.46\% & 1.23\% & 2.47\% & 81\\
                              & Simple RAG & 67.90\% & 29.63\% & 2.47\% & 0\% & 81\\
                              & Adv. RAG & 61.90\% & 27.38\% & 10.71\% & 0\% & 84\\
        \hline
    \end{tabular}
    \caption{Results of the evaluation of the thread's conformity with the community goals}
    \label{tab:スレッド目標適合性評価}
\end{table*}

\begin{table*}[t]
    \centering
    \small
    \begin{tabular}{|c|c||c|c|c|c|c|}
        \hline
        Scenario                  & Conditions    & Compatible & \begin{tabular}{c}Somewhat\\compatible\end{tabular} & \begin{tabular}{c}Somewhat\\incompatible\end{tabular} & Incompatible & Eval. total\\
        \hline
        \hline
        \multirow{3}{*}{\begin{tabular}{c}BOJ\\Discussion\end{tabular}} & Without RAG & 69.33\% & 22.67\% & 4.00\% & 4.00\% & 75\\
                              & Simple RAG & 65.38\% & 24.36\% & 6.41\% & 3.85\% & 78\\
                              & Adv. RAG & 67.90\% & 24.69\% & 7.41\% & 0\% & 81\\
        \hline
        \multirow{3}{*}{Otani Chat} & Without RAG & 70.37\% & 28.40\% & 1.23\% & 0\% & 81\\
                              & Simple RAG & 77.78\% & 20.99\% & 1.23\% & 0\% & 81\\
                              & Adv. RAG & 65.48\% & 22.62\% & 9.52\% & 2.38\% & 84\\
        \hline
    \end{tabular}
    \caption{Results of evaluation of threads for conformance to community rules}
    \label{tab:スレッドルール適合性評価}
\end{table*}

\begin{table*}[t]
    \centering
    \small
    \begin{tabular}{|c|c||c|c|c|c|c|}
        \hline
        Scenario                  & Conditions    & natural & somewhat natural & somewhat unnatural & unnatural & Eval. total\\
        \hline
        \hline
        \multirow{3}{*}{\begin{tabular}{c}BOJ\\Discussion\end{tabular}} & without RAG & 44.00\% & 25.33\% & 22.67\% & 8.00\% & 75\\
                              & Simple RAG & 33.33\% & 32.05\% & 17.95\% & 16.67\% & 78\\
                              & Adv. RAG & 59.26\% & 22.22\% & 17.28\% & 1.23\% & 81\\
        \hline
        \multirow{3}{*}{Otani Chat} & without RAG & 39.51\% & 39.51\% & 13.58\% & 7.41\% & 81\\
                              & Simple RAG & 33.33\% & 32.10\% & 24.69\% & 9.88\% & 81\\
                              & Adv. RAG & 44.05\% & 34.52\% & 17.86\% & 3.57\% & 84\\
        \hline
    \end{tabular}
    \caption{Results of the evaluation of the naturalness of the threads}
    \label{tab:スレッド自然さ評価}
\end{table*}

\subsection{Evaluation Results for Posts and Replies\label{sec:投稿評価}}

Tables \ref{tab:投稿目標適合性評価} through \ref{tab:投稿自然さ評価} show the results of the evaluation of posts and replies. Table \ref{tab:投稿目標適合性評価} shows the evaluation results of the conformity of posts and replies to the community goals, Table \ref{tab:投稿ルール適合性評価} shows the evaluation results of the conformity of posts and replies to the community rules, and The Table \ref{tab:投稿自然さ評価} shows the results of the evaluation of the naturalness of the overall post and reply texts. For reasons of space limitation, some results are presented in the Appendix.

In addition, the results of the evaluation of the textual integrity of the posts and replies are shown in Table \ref{tab:投稿文章評価} in Appendix \ref{sec:結果付録} and the results of the evaluation for persona suitability of posts or replies are shown in Table \ref{tab:投稿ペルソナ適合性評価} in Appendix \ref{sec:結果付録}.

The results show that, in general, the evaluations of the posts and replies generated by the proposed method are slightly inferior or equal to those of the other conditions. It can also be seen that while the posts generated by the simple RAG recorded the highest ratings in several evaluation items, the proposed method was only slightly below average in all of the evaluation items. 

This tendency is more pronounced in the evaluation of naturalness. As shown in Table \ref{tab:投稿自然さ評価}, in the ``Otani Chat'' scenario, the posts and replies generated by the proposed method have generally lower rating values than in the other conditions. On the other hand, the same scenario suggests that the simple RAG condition generated the most natural postings and replies.

\subsection{Evaluation Results for Threads\label{sec:スレッド評価}}

Tables \ref{tab:スレッド目標適合性評価} through \ref{tab:スレッド自然さ評価} show the results of the thread evaluation. Table \ref{tab:スレッド目標適合性評価} shows the evaluation results of the conformity of the thread contents to the community goals, Table \ref{tab:スレッドルール適合性評価} shows the evaluation results of the conformity of the thread contents to the community rule, and Table \ref{tab:スレッド自然さ評価} shows the results of the evaluation of the naturalness of the interactions within the thread. The number of evaluations is listed in ``Eval. total'' column, and the percentage of the evaluated value is listed in the other columns. For reasons of space limitation, some results are presented in the Appendix.

In addition, Table \ref{tab:スレッド文章評価} in Appendix \ref{sec:結果付録} shows the evaluation results for grammatical errors of the thread content.

From each table, it can be seen that, regardless of the scenario, the threads generated by the simple RAG condition obtain lower evaluation values than the other conditions. This tendency was more pronounced for the naturalness evaluation.

It can also be seen that the interaction generated by the proposed method has the highest average naturalness rating in both scenarios, although it is less compatible with the community interaction goals and community rules than the other conditions. It can also be seen that the threads generated by the proposed method recorded a maximum of 10\% higher evaluation values than the other conditions in terms of naturalness. This result contrasts with the results of the evaluation of individual posts and replies (see Table \ref{tab:投稿自然さ評価}), indicating that posts or replies that seem unnatural when evaluated alone may seem natural when they are part of an exchange.

\section{Consideration}

The results of the evaluation of postings and replies shown in Section \ref{sec:投稿評価} indicate that the postings and replies generated by the proposed method generally have lower evaluation values, although there are some differences depending on the scenario. This tendency is particularly noticeable in the ``Otani chat'' scenario, where, as an example, the number of times the rule conformity was evaluated as ``Compatible'' was 20\% lower than in the simple RAG mechanism. On the other hand, with respect to conformance to the exchange goals, rules, and personas, the no RAG condition shows stable conformance regardless of the scenario. These considerations suggest that conformity to exchange goals, rules, and personas may depend on the amount of information given by the RAG mechanism. It is possible that the large amount of information given by the RAG mechanism, i.e., the increased amount of references from articles in the prompt, makes the user persona and interaction goals/rules relatively unnoticed by the LLM, and that the posts/replies generated are also neglectful of these settings.The postings and replies generated are likely to neglect these settings.

In the SNS simulation system proposed in this paper, user characteristics are given as ``user personas'', which are short sentences of about several dozen characters, to give the LLMs the characteristics of the users they should role-play. In the experimental scenarios, the exchange goals and rules were given in relatively short sentences of about 20 characters. By describing these user persona and community goals and rules in more detail with a larger number of characters, it is expected that the generation will more faithfully follow the user persona and community goals and rules, even when the amount of information given by the RAG mechanism is large.

Overall, the ``Otani Chat'' scenario was rated higher than the ``Bank of Japan Discussion'' scenario under the no RAG and simple RAG mechanism conditions. On the other hand, the ``Bank of Japan Discussion'' scenario was rated higher than the ``Otani Chat'' scenario in the proposed method condition. This can be attributed to the nature of the Web sites selected as information sources. NHK NEWS WEB was used as the information source for the RAG mechanism in this system. While the site carries a wide range of news articles on politics and economics, the articles on Shohei Otani were mainly about his game results, which was considered inappropriate for the scenario. In order to perform high-quality simulations regardless of the scenario, it may be necessary for the simulation runner to provide information sources that can collect appropriate information for the scenario, or for the system to have a mechanism to actively select such information sources. This is an issue to be addressed in the future.

The results of the evaluation on threads presented in Section \ref{sec:スレッド評価}, as well as the results of the evaluation on posts, show that the proposed method condition is less compatible with the exchange goals and rules than the other conditions. This may be due to the low conformity of the individual contributions, which caused the exchange as a whole to deviate from the exchange goals and rules.

The evaluation of postings and replies alone suggested that the posts and replies generated by the proposed method were perceived as somewhat unnatural, while the evaluation of threads indicated that the threads generated by the proposed method were perceived as the most natural. In addition, the simple RAG mechanism condition, which also uses a RAG mechanism, shows that the exchange is perceived as rather more unnatural than the no RAG condition.

These suggest that it is possible to make the interaction more natural even when using the RAG mechanism by intentionally differentiating the amount of information available to individual chat agents. It was also suggested that providing uniform information to all chat agents may have a negative effect on conversation generation.

Overall, we can say that the proposed method can generate more natural postings, replies, and exchanges than other methods by using the RAG mechanism. This tendency is also noticeable for topics such as ``Bank of Japan Discussions'' that are suitable for information sources, and it can be said that the quality of SNS simulations using the proposed method is maximized when an appropriate information source is selected.

\section{Conclusion}

We propose a system that simulates interaction on a social networking service by giving LLM a community topic, rules, and user characteristics, and generating text-based postings between users in a large community. In this paper, we investigate how the RAG mechanism in our simulation system affects the quality and results of simulations, and conduct comparative experiments on simulation systems equipped with no RAG mechanism, a simple RAG mechanism, and a RAG mechanism of the proposed method, respectively. The results of the experiment confirmed that the proposed method, which reproduced individual differences in information gathering and reading comprehension, can generate the most natural exchange.

We also confirmed that a simple RAG mechanism that provides uniform information to all chat agents generates more conversations that are considered unnatural than the no RAG mechanism condition, in which no information is provided at all.

In general, we confirmed that the proposed method has excellent performance in SNS simulations, but we also confirmed that the simulation quality varies significantly depending strongly on the nature and characteristics of the information sources. In the future, it will be necessary to enable simulation of a wider range of scenarios through a mechanism that autonomously selects the appropriate one among multiple information sources.

\bibliography{myrefs}

\appendix

\section{Appendix\label{sec:appendix}}

\subsection{Appendix on The Proposed Method\label{sec:提案手法付録}}

This section contains an appendix on the proposed methodology. All Figures and Tables mentioned in this section are appendices.

First, Figure \ref{tab:ペルソナ生成プロンプト} in Appendix \ref{sec:提案手法付録} shows the prompts used to generate the user personas described in Section \ref{sec:ペルソナ生成}. The community goal set is inserted in ``\verb|{community goal}|'' in the prompt. Also in ``\verb|{sample user name}|'' and ``\verb|{sample user persona}|'', the user persona and name of the sample or generated user are inserted. These will be the same in subsequent prompts. 

The ``\verb|%%Repeat from here%%|'' to ``\verb|%%Repeat end%%|'' sections are repeated for the 10 selected sample users.

Figures \ref{tab:クエリ生成プロンプト投稿時} in Appendix \ref{sec:提案手法付録} and \ref{tab:クエリ生成プロンプト返信時} in Appendix \ref{sec:提案手法付録} show the prompts used to create search queries in Section \ref{sec:RAG}. Figure \ref{tab:クエリ生成プロンプト投稿時} in Appendix \ref{sec:提案手法付録} is the prompt used when generating a top-level posts. Figure \ref{tab:クエリ生成プロンプト返信時} in Appendix \ref{sec:提案手法付録} shows the prompt used when generating a reply. All posts in the targeted thread are inserted in \verb|{All posts in targeted thread}|in Figure \ref{tab:クエリ生成プロンプト返信時} in Appendix \ref{sec:提案手法付録}.

Table \ref{tab:要約文長さ} in Appendix \ref{sec:提案手法付録} shows how the RAG mechanism's summarization target and the length of the summarized text are set with the ``depth'' value. Title in the Target line refers to the title of the article, Abstruct refers to the summary text of the article content at the beginning of the article, and Main body refers to the main body of the article. The percentages in the table refer to the number of sentences after the summary relative to the number of sentences before the summary. As an example, if the depth value is 3 and the target article consists of 100 sentences, the number of sentences after summarization is 20.

Tables \ref{tab:投稿生成プロンプト} in Appendix \ref{sec:提案手法付録} and \ref{tab:返信生成プロンプト} in Appendix \ref{sec:提案手法付録} show the prompts used to generate the posts and replies described in Section \ref{sec:投稿生成}. In both prompts,  community rule are inserted in \verb|{community rules}| and information collected and summarized by the RAG mechanism is inserted in \verb|{Information}|. The name of the user who has already created a post or reply in the thread is inserted in \verb|{reply target}| in Table \ref{tab:返信生成プロンプト} in Appendix \ref{sec:提案手法付録}

\subsection{Appendix on Evaluation Results\label{sec:結果付録}}

This section contains an appendix on the evaluation results. All Figures and Tables mentioned in this section are appendices.

Table \ref{tab:投稿文章評価} in Appendix \ref{sec:結果付録} shows the results of the evaluation of the textual integrity of the posts. The evaluation results show that the system is able to generate posts with correct content in terms of sentence structure under all conditions.

Table \ref{tab:投稿ペルソナ適合性評価} in Appendix \ref{sec:結果付録} shows the results of having users evaluate the appropriateness of the posted content for a given user persona. It can be seen that the highest conformity was recorded in the condition without RAG, and slightly lower conformity was recorded in both conditions with RAG.

Table \ref{tab:スレッド文章評価} in Appendix \ref{sec:結果付録}  shows the results of the evaluation of textual consistency with respect to thread content. Unlike the evaluation results for individual posts shown in Table \ref{tab:投稿文章評価} in Appendix \ref{sec:結果付録}, the threads generated by the proposed method have the highest textual integrity.

\begin{figure*}[t]
    \centering
    \small
    \begin{tabular}{p{.9\linewidth}}
        \hline
        \verb|### Instructions|\\
        The input is the name and profile of the user participating in the net bulletin board for the purpose of "\verb|{community goal}|".\\
        Create a new user name and profile in the same format, using the input as a reference.\\
        Usernames should be no longer than 20 characters and profiles should be no longer than 30 characters.\\
        The user name and profile should be distinguishable from other users.Similar usernames and profiles are not allowed.\\
        \\
        \verb|### Input|\\
        \verb|%%Repeat from here%%|\\
        \verb|<pre class="user-name" max-length="20">|\\
        \verb|{sample user name}</pre> : |\\
        \verb|<pre class="user-description" max-length="30">|\\
        \verb|{sample user persona}</pre>|\\
        \verb|%%Repeat end%%|\\
        \verb|<pre class="name" max-length="20">|\\
        \hline
    \end{tabular}
    \caption{Prompt to generate user personas}
    \label{tab:ペルソナ生成プロンプト}
\end{figure*}

\begin{figure*}[t]
    \centering
    \small
    \begin{tabular}{p{.9\linewidth}}
        \hline
        \verb|### Instructions|\\
        You are \verb|{user name}|, \verb|{user persona}|.\\
        Role play the position of \verb|{user name}|, \verb|{user persona}|.\\
        Output the search words for the search on the input theme.\\
        Search words should be output in the form of an AND search, with multiple keywords connected by one-byte spaces.\\
        The search words should be separated by a new line for each session.\\
        \\
        \verb|### Input|\\
        \verb|{community goal}|\\
        \\
        \verb|### Response|\\
        \verb| - sample keyword|\\
        \verb| -|\\
        \hline
    \end{tabular}
    \caption{Prompt to generate search query when creating a top-level post}
    \label{tab:クエリ生成プロンプト投稿時}
\end{figure*}

\begin{figure*}[t]
    \centering
    \small
    \begin{tabular}{p{.9\linewidth}}
        \hline
        \verb|### Instructions|\\
        You are \verb|{user name}|, \verb|{user persona}|.\\
        Role play the position of \verb|{user name}|, \verb|{user persona}|.\\
        Reply to the input conversation. In doing so, output search words \\when you search the Web for information you need as a reference\\
        Search words should be output in the form of an AND search, with multiple keywords connected by one-byte spaces.\\
        The search words should be separated by a new line for each session.\\
        \\
        \verb|### Input|\\
        \verb|{All posts in targeted thread}|\\
        \\
        \verb|### Response|\\
        \verb| - sample keyword|\\
        \verb| -|\\
        \hline
    \end{tabular}
    \caption{Prompt to generate search query when creating a replies}
    \label{tab:クエリ生成プロンプト返信時}
\end{figure*}

\begin{table*}[t]
    \centering
    \small
    \begin{tabular}{|c||c|c|c|c|}
         \hline
         \multirow{2}{*}{value of ``depth''} & \multirow{2}{*}{Target} & \multicolumn{2}{c|}{Amount of indivisual sentences} & \multirow{2}{*}{\begin{tabular}{c}Amount of total sentences\end{tabular}}\\
         \cline{3-4}
         &  & $l<500$ & $l\geq500$ & \\
         \hline
         \hline
         0 & Title & 100\% & \multicolumn{2}{|c|}{100\%}\\
         1 & Abstruct & 100\% & \multicolumn{2}{|c|}{100\%}\\
         2 & Main body & 10\% & \multicolumn{2}{|c|}{5\%}\\
         3 & Main body  & 20\% & \multicolumn{2}{|c|}{10\%}\\
         4 & Main body  & 30\% & \multicolumn{2}{|c|}{15\%}\\
         5 & Main body  & 40\% & \multicolumn{2}{|c|}{20\%}\\
         6 & Main body  & 50\% & \multicolumn{2}{|c|}{25\%}\\
         \hline
    \end{tabular}
    \caption{Number of sentences after summarizing the original number of sentences}
    \label{tab:要約文長さ}
\end{table*}

\begin{figure*}[t]
    \centering
    \small
    \begin{tabular}{p{.9\linewidth}}
        \hline
        \verb|### Instructions|\\
        Responses must adhere to the rule "\verb|{community rule}|".\\
        Responses should be made in Japanese, role-playing the position of \verb|{user persona}|.\\
        The style of the response should be natural for a text posted on the Internet.\\
        If the information to be used is presented, it should be used to generate a response.\\
        \\
        \verb|### Available Information|\\
        \verb|{Information}|\\
        \\
        \verb|### Input|\\
        Consider the text you post on a bulletin board for the purpose of "\verb|{community goal}|" from the perspective of \verb|{user persona}|.\\
        \\
        \verb|### Response|\\
        \verb|[{user name}]:|\\
        \hline
    \end{tabular}
    \caption{Prompt to generate top-level post body}
    \label{tab:投稿生成プロンプト}
\end{figure*}

\begin{figure*}[t]
    \centering
    \small
    \begin{tabular}{p{.9\linewidth}}
        \hline
        \verb|### Instructions|\\
        Input is a message board thread intended for "\verb|{community goal}|".\\
        Role-play the user "\verb|{user name}|" as "\verb|{user persona}|" and generate a reply to the input thread in Japanese.\\
        The style of the response should be natural for a text posted on the Internet.\\
        If the information to be used is presented, it should be used to generate a response.\\
        \\
        \verb|### Available Information|\\
        \verb|{Information}|\\
        \\
        \verb|### Input|\\
        \verb|{thread}|\\
        \\
        \verb|### Response|\\
        \verb|[{user name}][ReplyTo: {reply target}]: [|\\
        \hline
    \end{tabular}
    \caption{Prompt to generate replies}
    \label{tab:返信生成プロンプト}
\end{figure*}

\begin{table*}[t]
    \centering
    \small
    \begin{tabular}{|c|c||c|c|c|c|}
        \hline
        Scenario                  & Conditions    & None at all & Almost none & Slightly & Many \\
        \hline
        \hline
        \multirow{3}{*}{BOJ Discussion} & Without RAG & 159 & 203 & 44 & 8\\
                              & Simple RAG & 199 & 166 & 44 & 5\\
                              & Adv. RAG & 186 & 170 & 50 & 8\\
        \hline
        \multirow{3}{*}{Otani Chat} & Without RAG & 195 & 161 & 55 & 3\\
                              & Simple RAG & 151 & 169 & 83 & 11\\
                              & Adv. RAG & 199 & 145 & 66 & 4\\
        \hline
    \end{tabular}
    \caption{Results of evaluation of grammatical errors of posts and replies}
    \label{tab:投稿文章評価}
\end{table*}

\begin{table*}[t]
    \centering
    \small
    \begin{tabular}{|c|c||c|c|c|c|}
        \hline
        Scenario                  & Conditions    & Compatible & Somewhat compatible & Somewhat incompatible & Incompatible \\
        \hline
        \hline
        \multirow{3}{*}{BOJ Discussion} & Without RAG & 217 & 144 & 41 & 12\\
                              & Simple RAG & 191 & 140 & 68 & 15\\
                              & Adv. RAG & 202 & 141 & 57 & 14\\
        \hline
        \multirow{3}{*}{Otani Chat} & Without RAG & 216 & 125 & 38 & 35\\
                              & Simple RAG & 223 & 132 & 45 & 14\\
                              & Adv. RAG & 192 & 129 & 66 & 27\\
        \hline
    \end{tabular}
    \caption{Results of evaluation of persona suitability of posts and replies}
    \label{tab:投稿ペルソナ適合性評価}
\end{table*}

\begin{table*}[t]
    \centering
    \small
    \begin{tabular}{|c|c||c|c|c|c|c|}
        \hline
        Scenario                  & Conditions    & None at all & Almost none & slightly & many & Eval. total\\
        \hline
        \hline
        \multirow{3}{*}{\begin{tabular}{c}BOJ\\Discussion\end{tabular}} & Without RAG & 25.33\% & 52.00\% & 20.00\% & 2.67\% & 75\\
                              & Simple RAG & 12.82\% & 62.82\% & 17.95\% & 6.41\% & 78\\
                              & Adv. RAG & 25.93\% & 50.62\% & 19.75\% & 3.70\% & 81\\
        \hline
        \multirow{3}{*}{Otani Chat} & Without RAG & 11.11\% & 58.02\% & 22.22\% & 8.64\% & 81\\
                              & Simple RAG & 19.75\% & 40.74\% & 30.86\% & 8.64\% & 81\\
                              & Adv. RAG & 28.57\% & 46.43\% & 22.62\% & 2.38\% & 84\\
        \hline
    \end{tabular}
    \caption{Results of the evaluation of the grammatical errors of the thread}
    \label{tab:スレッド文章評価}
\end{table*}

\end{document}